\documentclass[sigconf]{acmart}

\AtBeginDocument{
  }

\acmSubmissionID{2189}

\usepackage{colortbl}
\usepackage[capitalize]{cleveref}
\crefname{section}{Sec.}{Secs.}
\Crefname{section}{Section}{Sections}
\Crefname{table}{Table}{Tables}
\crefname{table}{Tab.}{Tabs.}

\copyrightyear{2023} 
\acmYear{2023} 
\setcopyright{acmlicensed}\acmConference[MM '23]{Proceedings of the 31st ACM International Conference on Multimedia}{October 29-November 3, 2023}{Ottawa, ON, Canada}
\acmBooktitle{Proceedings of the 31st ACM International Conference on Multimedia (MM '23), October 29-November 3, 2023, Ottawa, ON, Canada}
\acmPrice{15.00}
\acmDOI{10.1145/3581783.3612192}
\acmISBN{979-8-4007-0108-5/23/10}
\begin{document}

\title[On the Importance of Spatial Relations for Few-shot Action Recognition]{On the Importance of Spatial Relations for Few-shot \\ Action Recognition}

\settopmatter{authorsperrow=4}

\author{Yilun Zhang$^{\ast}$}
\orcid{0009-0001-8471-6983}
\affiliation{
  \institution{Shanghai Key Lab of Intel. Info. Proc., School of CS, Fudan University}
  \country{}
  }
  
\author{Yuqian Fu$^{\ast}$}
\affiliation{
  \institution{Shanghai Key Lab of Intel. Info. Proc., School of CS, Fudan University}
  \country{}
  }

\author{Xingjun Ma$^\#$}
\affiliation{
  \institution{Shanghai Key Lab of Intel. Info. Proc., School of CS, Fudan University}
  \country{}
  }

\author{Lizhe Qi}
\affiliation{
  \institution{Academy for Engineering and Technology, Fudan University}
  \country{}
  }

\author{Jingjing Chen}
\affiliation{
  \institution{Shanghai Key Lab of Intel. Info. Proc., School of CS, Fudan University}
  \country{}
  }

\author{Zuxuan Wu}
\affiliation{
  \institution{Shanghai Key Lab of Intel. Info. Proc., School of CS, Fudan University}
  \country{}
  }

\author{Yu-Gang Jiang$^\#$}
\affiliation{
  \institution{Shanghai Key Lab of Intel. Info. Proc., School of CS, Fudan University}
  \country{}
  }

\thanks{$\ast$ indicates equal contribution, $\#$ indicates corresponding authors. }

\renewcommand{\shortauthors}{Zhang et al.}

\begin{abstract}
Deep learning has achieved great success in video recognition, yet still struggles to recognize novel actions when faced with only a few examples. To tackle this challenge, few-shot action recognition methods have been proposed to transfer knowledge from a source dataset to a novel target dataset with only one or a few labeled videos. 
However, existing methods mainly focus on modeling the temporal relations between the query and support videos while ignoring the spatial relations.
In this paper, we find that the spatial misalignment between objects also occurs in videos, notably more common than the temporal inconsistency.
We are thus motivated to investigate the importance of spatial relations and propose a more accurate few-shot action recognition method that leverages both spatial and temporal information.
Particularly, a novel Spatial Alignment Cross Transformer (SA-CT) which learns to re-adjust the spatial relations and incorporates the temporal information is contributed.
Experiments reveal that, even without using any temporal information, the performance of SA-CT is comparable to temporal based methods on 3/4 benchmarks.
To further incorporate the temporal information, we propose a simple yet effective Temporal Mixer module.
The Temporal Mixer enhances the video representation and improves the performance of the full SA-CT model, achieving very competitive results.
In this work, we also exploit large-scale pretrained models for few-shot action recognition, providing useful insights for this research direction.
\end{abstract}



\begin{teaserfigure}
  \includegraphics[width=\textwidth]{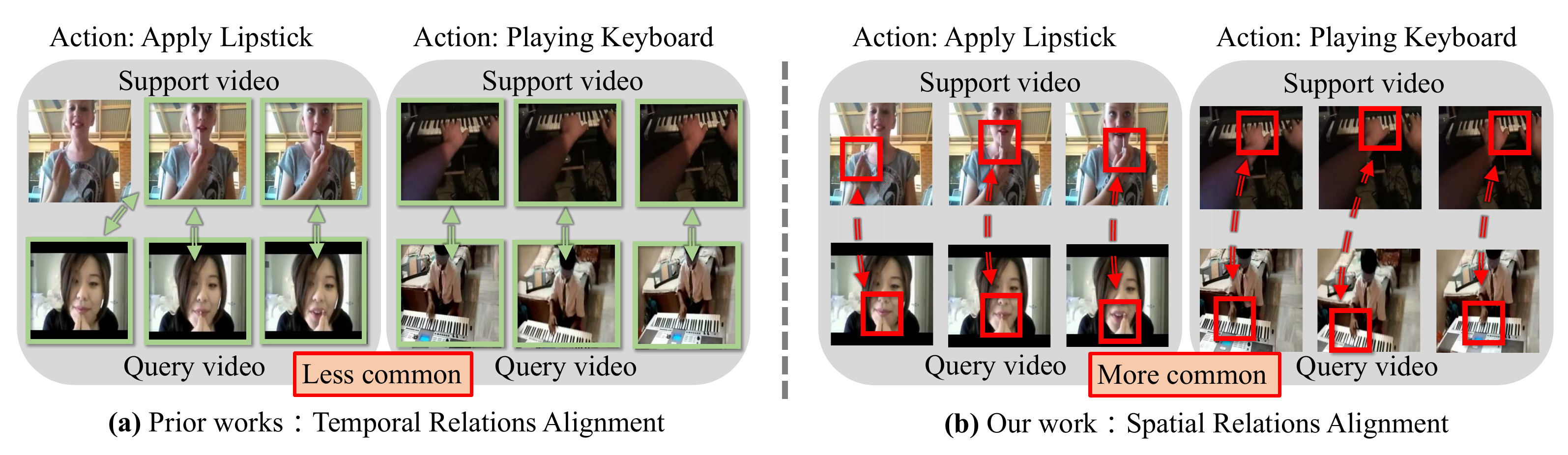}
    \vspace{-0.3in}
  \caption{\textbf{Temporal relations alignment (prior works) vs. spatial relations alignment (our work).}
    Compared to temporal inconsistency in (a), the spatial misalignment issue in (b) is more common. }
  \label{fig:intro}
\end{teaserfigure}

\maketitle

\section{Introduction}
The massive amount and ever-growing video data demand automated video recognition techniques to analyze video content effectively for a multitude of multimedia applications.
Powered by deep neural networks (DNNs), the performance of this task has been greatly improved in the past few years~\cite{feichtenhofer2019slowfast, tran2015learning, carreira2017quo, qiu2017learning, lin2019tsm}. 
However, most of the mainstream video recognition methods rely heavily on the availability of massive labeled training data for improved performance. This is to assume that each target action has a large number of labeled examples. Unfortunately, such an assumption does not always hold in real-world applications. For example, it is unrealistic to collect videos for the actions of \textit{drowning}, \textit{failing}, etc. 
Recognizing actions in such a low-sample regime is thus of great importance---a challenging task also known as \emph{few-shot action recognition} (FSAR).

Given a novel action with only a few labeled examples (also called \emph{support videos}), the goal of FSAR is to classify the unseen query video according to the support videos. 
A body of work has been proposed to achieve accurate FASR from different perspectives~\cite{zhu2018compound, bishay2019tarn, fu2020depth, cao2020few, perrett2021temporal, thatipelli2022spatio, wu2022motion, wang2022hybrid}, amongst which exploiting the temporal relations between the query and the support videos is the latest and most effective approach. Representative methods include TAM~\cite{cao2020few}, MTFAN~\cite{wu2022motion}, TRX~\cite{perrett2021temporal}, and STRM~\cite{thatipelli2022spatio}. 
They all emphasize the importance of temporal relations and aim to match the support and query videos by aligning the temporal frames.
However, in this paper, as depicted in ~\cref{fig:intro}, we reveal that the spatial misalignment of key objects also occurs in FSAR and is even more common than the temporal inconsistency, an important observation that has been neglected in previous works.
This motivates us to re-examine the importance of spatial relations for few-shot action recognition, by answering the two questions: 1) \emph{Could spatial relations alone be sufficient for recognizing few-shot actions?} and 2) \emph{Can temporal information be simply utilized as a boost to the final video representations?}

To answer the first question, we propose a novel spatial cross-attention (SCA) module to model the spatial relations between the query and support videos. Specifically, we first split the whole video frames into patches of equal size and then learn the similarities between the support and query patches via cross-attention.
By matching those patches, objects of interest can be aligned even if they appear at different spatial locations.
Through the aligned spatial relations, we turn the original support features into query-specific support features, eliminating the negative effects caused by the spatial misalignment.
Experimentally, we find that our SCA module alone without integrating any temporal information is already very competitive. 
On three out of four benchmarks, the SCA module achieves comparable or even better results than the current SOTAs. This indicates that the spatial relations play a central role in FSAR.

For the second question, we explore an alternative approach to exploit the temporal information, i.e., using it to boost the representations rather than temporally (and expensively) aligning the query and support videos as it did in prior works.
Instead of designing sophisticated algorithms to match the temporal frames~\cite{bishay2019tarn, cao2020few}, we explore a simple Temporal Mixer (TMixer) module for integrating the temporal information. We build the TMixer with several MLPs which is quite simple. 

With the SCA module and the TMixer module, we are able to build a new state-of-the-art for FSAR.

Formally, a novel \textbf{S}patial \textbf{A}lignment \textbf{C}ross \textbf{T}ransformer (\textbf{SA-CT}) is proposed for few-shot action recognition. Our SA-CT is mainly composed of a feature extractor, a spatial cross-attention (SCA) module, and a Temporal Mixer (TMixer) module.
Given a query video and a set of support videos, a feature extractor is first applied to extract the frame patch representations. The TMixer module is then employed to enrich the representations. The SCA module matches the patch representations between the query and support to construct query-specific support representations for classification. 
With SA-CT and its modules, we reveal that: 1) modeling the spatial relations alone can achieve comparable or even better results in most cases; 2) the temporal information can be effectively exploited as a temporal boost to the representations via a mixer module.

In addition, inspired by the recent success of large-scale pretrained models (LSPMs), we take a step further to explore several popular LSPMs including CLIP~\cite{radford2021learning}, DINO~\cite{caron2021emerging}, and DeiT~\cite{touvron2021training} for FSAR. Concretely, we leverage the backbones of these models as feature extractors to extract more enriched representations for videos.
We empirically verify the benefit of using LSPMs for FSAR along with several useful insights. 
These explorations and analyses could help the community build more accurate FSAR models.

To summarize, we make the following contributions.
\textbf{1)} We reveal the importance of spatial relations for FSAR, a long-overlooked aspect in the literature, and propose a novel spatial cross-attention (SCA) module to model the spatial relations for more accurate FSAR. \textbf{2)} We propose to use a Temporal Mixer (TMixer) module to integrate the temporal information into the representations as a complementary and boost to the spatial information. \textbf{3)} We combine the SCA and TMixer module into a unified Spatial Alignment Cross Transformer (SA-CT) architecture which achieves very competitive results for FSAR. \textbf{4)} We also provide the extensive exploration of LSPMs as more powerful feature extractors for FSAR.

\section{Related Work}
\label{sec:Related}

\noindent\textbf{Few-Shot Learning (FSL).} 
Typical FSL methods can be roughly divided into three categories: model-based~\cite{santoro2016meta, munkhdalai2017meta}, metric-based~\cite{koch2015siamese, vinyals2016matching, snell2017prototypical, sung2018learning} and optimization-based~\cite{ravi2016optimization, finn2017model}.
Model-based methods aim to quickly update the parameters on a small number of samples through the design of the model structure, and directly establish the mapping function between the input instances and the predictions.
Metric-based methods measure the distances (e.g., cosine similarity) between the samples in the support set and query set.
More recent FSL methods~\cite{zhmoginov2022hypertransformer, fu2021meta, tang2020blockmix, doersch2020crosstransformers, zhang2021shallow, tang2022learning,  zhang2022progressive, zhuo2022tgdm, fu2023styleadv, zhang2023deta} include HyperTransformer~\cite{zhmoginov2022hypertransformer}, Meta-FDMixup~\cite{fu2021meta}, STraTA~\cite{vu2021strata}, CrossTransformers~\cite{doersch2020crosstransformers}, MetaQDA~\cite{zhang2021shallow}, PMF~\cite{hu2022pushing}, and StyleAdv~\cite{fu2023styleadv}.
Among them, CrossTransformers~\cite{doersch2020crosstransformers} is the most related work to ours, which uses the attention mechanism to find spatial correspondence between the query and the labeled images. However, CrossTransformers was proposed for image few-shot learning, while we tackles the more challenging problem of video few-shot learning.

\noindent\textbf{Few-Shot Action Recognition (FSAR).} FSAR aims to recognize unseen videos with only a few labeled samples. 
Prior works have made certain progresses via including compound memory network~\cite{zhu2018compound} for optimal video representations, synthesizing additional examples for novel categories~\cite{kumar2019protogan}, leveraging synthetic videos as data augmentation~\cite{fu2019embodied}, and introducing extra multimodal information~\cite{fu2020depth}.
More recent works focus on utilizing temporal information. Inspired by the text sequence matching task, TARN~\cite{bishay2019tarn} regards videos as segment-level sequence data and matches the query with the support videos. OTAM~\cite{cao2020few} aligns query and support videos temporally by calculating frame similarities. In order to align sub-sequences of actions at different speeds, TRX~\cite{perrett2021temporal} constructs video representations from ordered tuples of varying numbers of frames. 
MTFAN~\cite{wu2022motion} explored the task-specific motion modulation and the multi-level temporal fragment alignment.
HyRSM~\cite{wang2022hybrid} brought up hybrid relation module and set matching metric.
By adding spatial and temporal enrichment module to TRX~\cite{perrett2021temporal}, STRM~\cite{thatipelli2022spatio} achieves impressive performance in FSAR.

\noindent\textbf{Cross-Attention for Few-Shot Learning.}
One core challenge of few-shot learning lies in matching the support and query instances. Following this, several methods~\cite{hou2019cross, doersch2020crosstransformers, perrett2021temporal, thatipelli2022spatio, zhao2022semantic} have been proposed to explore cross-attention for improved alignment of the image/video instances.
Among them, TRX~\cite{perrett2021temporal} and STRM~\cite{thatipelli2022spatio} are the two most related works to us.
Specifically, TRX shows its potential in matching actions at different speeds. Based on TRX, STRM reaches very competitive performance by further adding a self-attention module on spatial patchs.
Different from TRX and STRM which both tackle FSAR from a temporal perspective, our work investigates the importance of spatial relations to FSAR and proposes to align the spatial objects between videos to achieve more accurate FSAR.
Note that STRM also handles the spatial relation, but via self-attention applied on patches within a single video. By contrast, we apply cross-attention to interact patches between the support and query videos.

\noindent\textbf{Large-Scale Pretrained Models (LSPMs).} Previous works have shown the significant impacts of LSPMs to various downstream tasks~\cite{dosovitskiy2020image, radford2021learning, liu2021swin, touvron2021training, caron2021emerging}. Those LSPMs exceed the CNN network in a number of vision tasks. A recent work~\cite{hu2022pushing} shows that a simple transformer-based pipeline can boost the performance of FSL. 
However, applying LSPMs for FSAR is still underexplored.
Existing FSAR methods~\cite{perrett2021temporal, thatipelli2022spatio, fu2020depth, cao2020few} mainly focus on a few standard CNN architectures (e.g., ResNet-50 pretrained on ImageNet). 
In this paper, we explore whether FSAR can also benefit from LSPMs and to what extent LSPMs can boost the performance of our model.

\section{Proposed Method}
\label{sec:Method}

\subsection{Problem Formulation}
\label{subsec:Problem Formulation}
Given $C$ action classes each containing only $K$ (a small number like 5) labelled instances as the ‘support set’, the task of FSAR is to classify an unlabelled query video into one of the classes in the ‘support set’. 
Following the episodic training paradigm in prior works ~\cite{vinyals2016matching, finn2017model, cao2020few, zhang2020few}, we use episodic training in which few-shot tasks are sampled randomly from the training set. In each episode, we learn a $C$-way $K$-shot classification task. We denote a query video of $L$ frames as $Q=\{q_1,\cdots ,q_L\}$. 
For each class $c \in \{1,\cdots,C\}$, we denote the support videos of this class as $S^c$. Specifically, $S^c$ contains $K$ videos, for the $k^{th}$ video, we have $S^c_k=\{s^c_{k1},\cdots ,s^c_{kL}\}$.

     \vspace{-0.1in}

\subsection{Method Overview}
\label{subsec:Overview}

\begin{figure*}
    \centering
    \includegraphics[width=1.\linewidth]{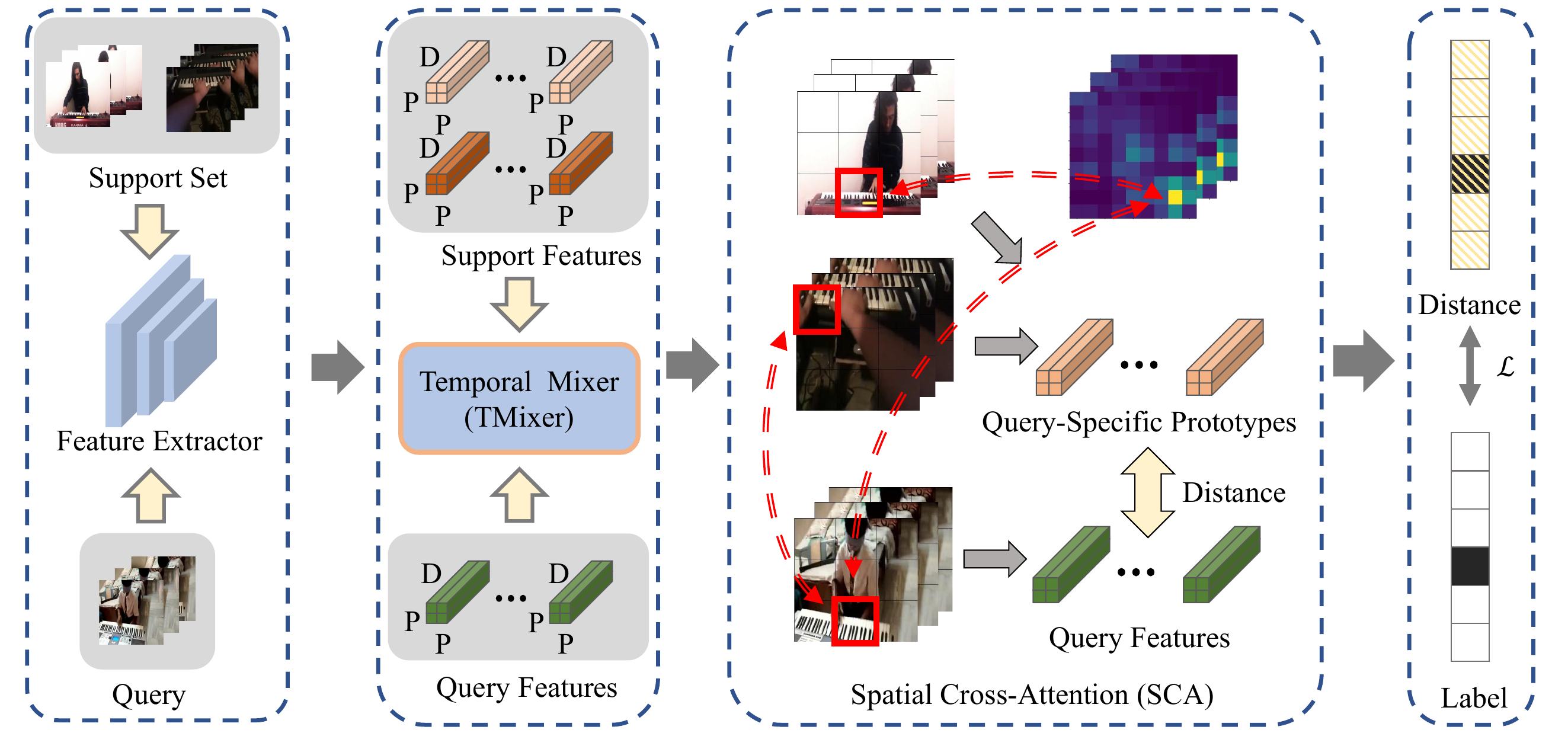}
    \caption{\textbf{The architecture of our proposed Spatial Alignment Cross Transformer (SA-CT), illustrated in a 1-way 2-shot case.} It consists of a 1) feature extractor that extracts the features for both the support set and the query; 2) a TMixer module that integrates the higher-order temporal information; and 3) a SCA module that aligns the spatial patches between the query and support set, and construct the query-specific prototypes for classification. 
    }
     \vspace{-0.15in}
  \label{fig:main}
\end{figure*}

Motivated by the observation that the spatial misalignment of key objects is oftentimes more severe than the temporal inconsistency, we introduce a novel \textbf{S}patial \textbf{A}lignment \textbf{C}ross \textbf{T}ransformer (\textbf{SA-CT}) to better handle the spatial misalignment by readjusting the spatial relations between two videos. 
The overall architecture is shown in ~\cref{fig:main}. First, a feature extractor is leveraged to encode the video frames in the support and query sets. The frame features  are then passed through the TMixer module to fuse higher-order temporal information as well as to reduce the number of video frames. Next, the SCA module is employed to match the spatial patches between the query and support set, and construct the query-specific prototypes for classification. Finally, distances are passed as logits for training and inference steps.

\begin{figure}
    \centering
    \includegraphics[width=.85\linewidth]{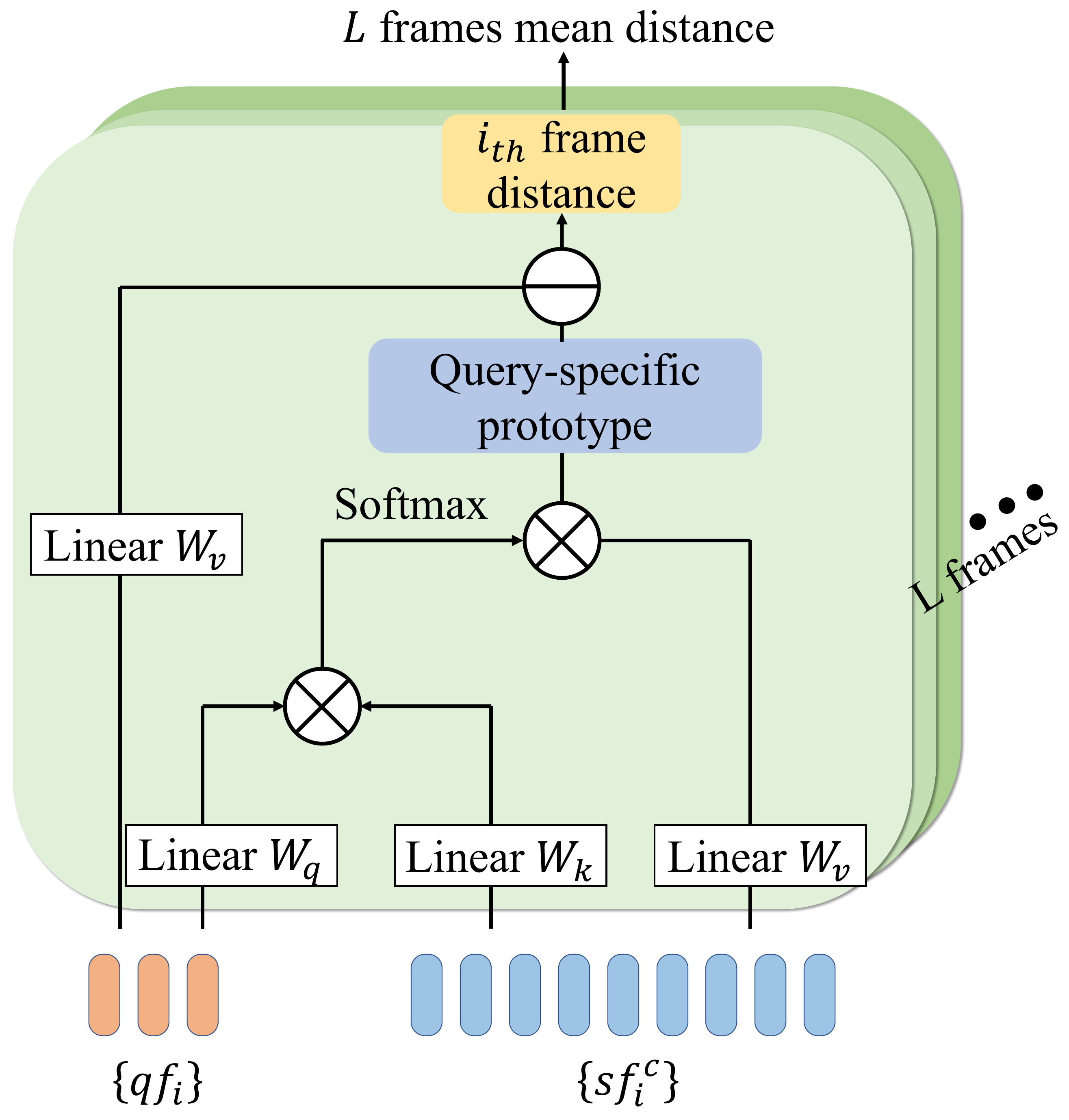}
             \vspace{-0.15in}

    \caption{\textbf{The architecture of the SCA module}. Each corresponding frame of patches in the query and support videos are first passed through the Linear weights to compute their cross attention. A query-specific prototype is then constructed based on the attention map. Finally, a mean distance value is computed by averaging over the distances of all video frames.}
     \vspace{-0.15in}
  \label{fig:spatial_crossattention}
\end{figure}

\subsection{Spatial Cross-Attention Module (SCA)}
\label{subsec:spatial cross attention}

Cross-attention mechanism, initially introduced in ~\cite{vaswani2017attention}, has shown its ability in aligning images and videos~\cite{perrett2021temporal, thatipelli2022spatio, fu2020depth, cao2020few}. We employ this mechanism in our SA-CT to align the spatial objects in the query and support set. The procedure is illustrated in ~\cref{fig:spatial_crossattention}. Specifically, we define the query feature of the $i^{th}$ ($i \in[1, L]$) frame at the spatial position $p$ as:
\begin{equation}
    qf_{i p}=\left[\Phi\left(q_i\right)_p+\operatorname{CPE}\left(\Phi\left(q_i\right)_p\right)\right], 
\end{equation}
where $\Phi: \mathbb{R}^{H \times W \times 3} \mapsto \mathbb{R}^{P^2 \times D}$ represents a feature extractor that extracts the latent features of $P^2$ patches (i.e., spatial position $p \in [0, P^2]$), and $\operatorname{CPE}(\cdot)$ is a conditional positional encoding of a frame feature~\cite{chu2021conditional}.
Similarly, the feature at spatial position $m$ of the $i^{th}$ ($i \in[1, L]$) frame of video $k$ in the support set of class $c$ is:
\begin{equation}
    sf_{i k m}^c=\left[\Phi\left(s_{i k}^c\right)_m+\operatorname{CPE}\left(\Phi\left(s_{i k}^c\right)_m\right)\right].
\end{equation}

After generating the features of the query and support videos, the similarity between the query and support at the $i^{th}$ ($i \in[1, L]$) frame can be calculated as:
\begin{equation}
    a_{i k m p}^c=LN\left(W_q \cdot sf_{i k m}^c\right) \cdot LN\left(W_k \cdot qf_{i p}\right),
\end{equation}
where $LN(\cdot)$ is the standard layer normalization~\cite{ba2016layer}, and $W_q$, $W_k$ are learnable projections of the features into the query and key embeddings used in the attention mechanism~\cite{vaswani2017attention}.

 For spatial alignment, each of the support videos is fully utilized, i.e., 
 patches in the query video are matched with those at different locations of all the videos in support class $c$. 
 The attention map can thus be derived by applying the Softmax operation along patches of videos in a support class:
\begin{equation}\label{eq:attention_map}
    \tilde{a}_{i k m p}^c=\frac{\exp \left(a_{i k m p}^c / \tau\right)}{\sum_{l, n} \exp \left(a_{i l n p}^c / \tau\right)} , \tau=\sqrt{d_k}.
\end{equation}

The above attention map represents the correspondences between different spatial locations in the query and support videos. This allows the module to readjust the support features into a query-specific prototype for better spatial alignment as follows:
\begin{equation}
    \mathbf{t}_{i p}^c=\sum_{k m} \tilde{a}_{i k m p}^c \cdot \left(\mathbf{W_v}\cdot \Phi\left(s_{i k}^c\right)_m\right),
\end{equation}
where, $\mathbf{W_v}$ represents the attention value weights.

Finally, the module calculates the distances between the query-specific prototypes and the query using squared Euclidean distance, and parses the distances as logits to represent the distribution over the classes:
\begin{equation}
d\left(Q, S^c\right)=\frac{1}{P^2} \sum_p\left\|\frac{1}{L^2} \sum_i\left(\boldsymbol{t}_{i p}^c-\left(\mathbf{W_v}\cdot \Phi\left(q_i \right)_p\right)\right)\right\|_2^2.
\end{equation}

 The same value weight $W_v$ is applied to both the query and support features to ensure that the distances calculated above could measure the similarities between the query and support videos. We also set the query weight $W_q$ and key weight $w_k$ to be the same, which could maximize the attention value for the corresponding spatial locations.

     \vspace{-0.1in}

\subsection{Temporal Mixer Module (TMixer)}
\label{subsec:Temporal mixer}

\begin{figure}
    \centering
    \includegraphics[width=1.\columnwidth]{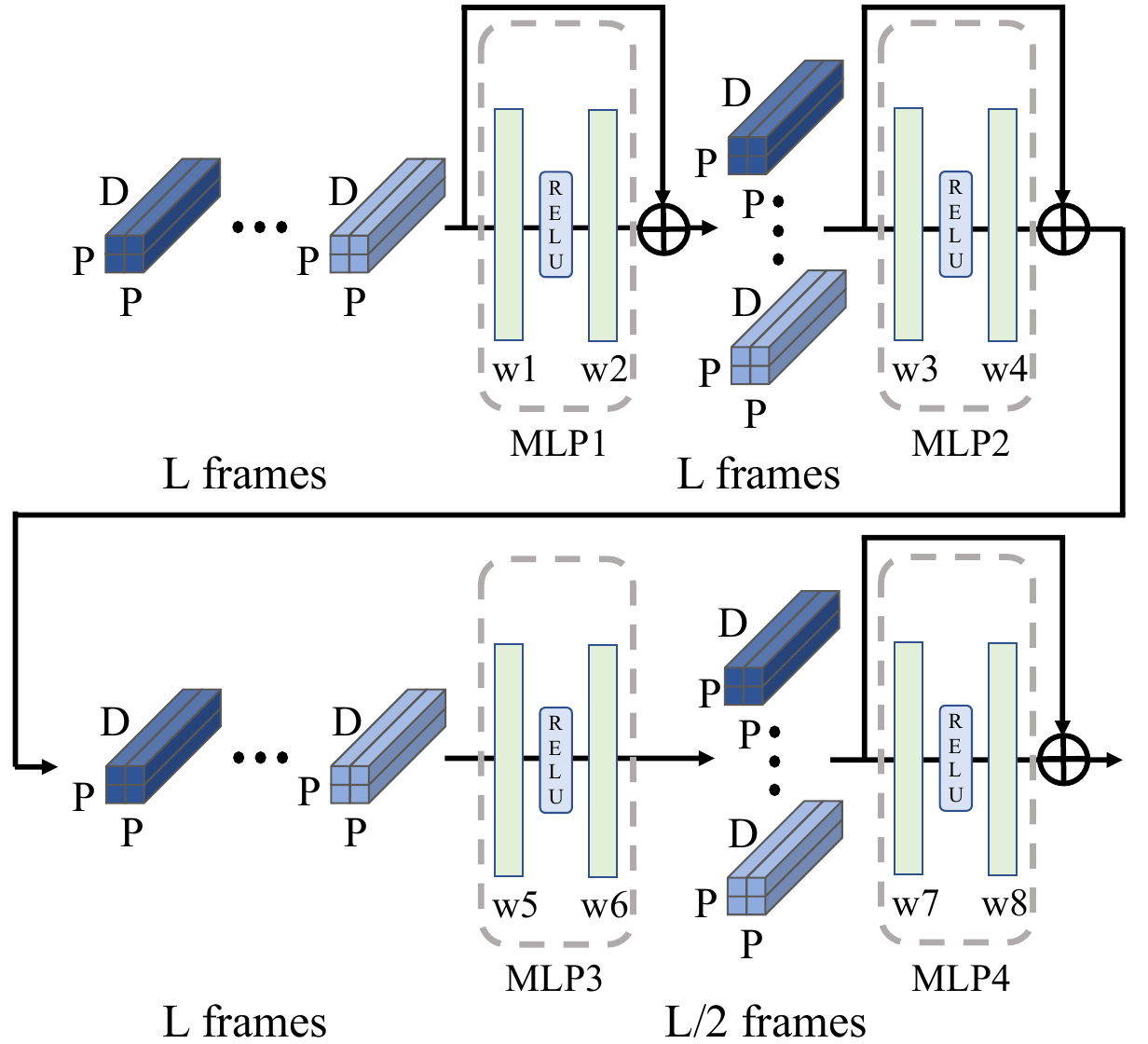}
         \vspace{-0.15in}

    \caption{\textbf{The architecture of the TMixer module}. The frame representations are passed through four MLPs, each of which consists two layers. 
    The first two MLPs integrate the temporal information by allowing the frames to interact with each other.
    The last two MLPs are employed to reduce the number of frames, thus decreasing the computational cost and accelerating the model.
    }
    \vspace{-0.25in}
  \label{fig:temporal_mixer}
\end{figure}

The SCA module enables our SA-CT to align  objects in the query and support videos. In addition to spatial relations, we explore a TMixer module that modifies two MLP-mixer~\cite{tolstikhin2021mlp} layers to integrate the temporal information in a more efficient manner.
A standard MLP-Mixer consists of two types of MLP layers: channel-mixing MLPs and token-mixing MLPs. 
The main purpose of channel-mixing MLPs is to facilitate communication between different channels, which inspires us to integrate the inter-frame relations with a global reception field.
In our TMixer module, we regard frames of a video as different channels. Frame features are thus able to interact with each other and be enriched globally with high-order temporal information. An illustration of this module is depicted in ~\cref{fig:temporal_mixer}. Concretely, consider the feature $f_i$ obtained by a backbone at the $i_{th}$ ($i \in[1, L]$) frame, then the concatenated feature representations of an entire video are denoted as $\mathbf{F}=\left[\mathbf{f}_1 ; \cdots ; \mathbf{f}_L\right] \in \mathbb{R}^{L \times P^2 \times D}$. We employ two MLPs, to interact frames of a single video:
\begin{equation}
    \mathbf{U}_{i , * , *}=\mathbf{F}_{i , * , *}+\mathbf{W}_2 \cdot \sigma\left(\mathbf{W}_1 \cdot \mathbf{F}_{i , * , *}\right), 
    i=1 \ldots L,
\end{equation}
\begin{equation}
    \mathbf{V}_{* , * , j}=\mathbf{U}_{* , * , j}+\mathbf{W}_4 \cdot \sigma\left(\mathbf{W}_3 \cdot \mathbf{U}_{* , * , j}\right),  
    j=1 \ldots D,
\end{equation}
where, $\sigma$ denotes the ReLU non-linearity, and $\mathbf{W}_1$, $\mathbf{W}_2$ $\in \mathbb{R}^{L\times L}$, $\mathbf{W}_3$, $\mathbf{W}_4$ $\in \mathbb{R}^{D\times D}$ are all learnable weights.

We have enriched the video features globally following the idea of frame mixing, and now we further extend this idea to reducing the frames.
Concretely, the following two MLPs are used to reduce the frames and accelerate the aforementioned SCA modules:
\begin{equation}
    \mathbf{Y}_{* , * , *}=\mathbf{W}_6 \cdot \sigma\left(\mathbf{W}_5 \cdot \mathbf{V}_{i , * , *}\right),  
    i=1 \ldots L,
\end{equation}
\begin{equation}
    \mathbf{Z}_{* , * , j}=\mathbf{Y}_{* , * , j}+\mathbf{W}_8 \cdot \sigma\left(\mathbf{W}_7 \cdot \mathbf{Y}_{* , * , j}\right),  
    j=1 \ldots D,
\end{equation}
where, $\sigma$ denotes the ReLU non-linearity, and $\mathbf{W}_5$ $\in \mathbb{R}^{L\times L/2}$ $\mathbf{W}_6$ $\in \mathbb{R}^{L/2\times L/2}$, $\mathbf{W}_7$, $\mathbf{W}_8$ $\in \mathbb{R}^{D\times D}$ are all learnable weights.
After passing through these two MLPs, $L$ frames in each video are reduced to $L/2$ frames, we will empirically verify the effectiveness of this operation in ~\cref{subsec:Ablation Study}.

Note that this module is applied on both the query and the support features before passing through the SCA module. As such, features of one single frame are enabled to incorporate high-order temporal information of the whole video.

\subsection{Relation to Prior Works}
\label{subsec:relation}

The recent works TRX~\cite{perrett2021temporal} and STRM~\cite{thatipelli2022spatio} both explore using cross-attention to address FSAR tasks. TRX~\cite{perrett2021temporal} focuses on aligning sub-sequences of videos. On the other hand, STRM~\cite{thatipelli2022spatio} incorporates spatial information by employing self-attention on individual video frames, enriching them locally, and then passing these enriched frames into a modified TRX module (named TRM) for recognition. Despite this, the core concept of STRM remains similar to TRX, as it still involves video sub-sequences matching.
In contrast, our approach differs from TRX and STRM. We utilize cross-attention to match patches in both support and query video frames, enabling an interaction between all the support and query patches. This enables our proposed SA-CT model to spatially align videos effectively.

\section{Experiments}
\label{sec:Experiments}
\subsection{Experimental Setup}
\label{subsec:Setup}
\paragraph{\textbf{Datasets.}} 

We evaluate our method on four action recognition datasets, including HMDB51~\cite{kuehne2011hmdb}, UCF101~\cite{soomro2012ucf101}, Kinetics~\cite{carreira2017quo}, and Something-Something V2 (SSv2)~\cite{goyal2017something}.
UCF101 dataset contains 101 action categories of over 13,320 short trimmed videos; HMDB51 dataset has 6,849 videos from 51 action categories; Kinetics contains 400 types of human actions and each video lasts around 10 seconds; SSv2 is the largest dataset containing over 220k videos.
One common characteristic of the UCF101, HMDB51, and Kinetics datasets is that the semantic concepts (e.g., objects and backgrounds) are more related to the action categories.
However, for SSv2, a large proportion of the categories are more related to the temporal information~\cite{sevilla2021only}.
As for the specific splits of training, validation, and testing sets, for UCF101 and HMDB51, we use the splits as in ARN~\cite{zhang2020few}. For Kinetics, we follow the splits proposed in CMN~\cite{zhu2018compound}, in which 100 classes are selected to split into 64/12/24 classes for train/val/test, respectively. 
For SSv2, we follow the same splits as in OTAM~\cite{cao2020few}, which also define 64/12/24 action classes for train/val/test sets.

\begin{table*}
    \centering

    \setlength{\tabcolsep}{2.9mm}{
    \begin{tabular*}{1\linewidth}{@{}cccccccccc@{}}
    \toprule 
    \textbf{Method} &  \textbf{Backbone} & \multicolumn{2}{c}{\textbf{UCF101}} & \multicolumn{2}{c}{\textbf{HMDB51}} & \multicolumn{2}{c}{\textbf{SSv2}} & \multicolumn{2}{c}{\textbf{Kinetics}} \\
     \cmidrule{3-10} 
    &  & 1 shot & 5 shot & 1 shot & 5 shot & 1 shot & 5 shot & 1 shot & 5 shot\\
    \midrule 
    CMN~\cite{zhu2018compound} & ResNet-50 & - & - & - & - & - & - &60.5 & 78.9 \\
    
    TARN~\cite{bishay2019tarn} & ResNet-50 & -  & - & - & - & - & - & 64.8 & 78.5  \\
    Embodied~\cite{fu2019embodied} & ResNet-50 & - & - & - & - & - & - & 67.8 & 85.0 \\
    ARN~\cite{zhang2020few} & ResNet-50 & 66.3 & 83.1 & 45.5 & 60.6& - & - & 63.7& 82.4 \\

    OTAM~\cite{cao2020few} & ResNet-50 & - & - & - & - & 42.8 & 52.3 & 73.0 &85.8 \\
   
    AmeFu-Net~\cite{fu2020depth} & ResNet-50 & 85.1 & 95.5 & 60.2 & 75.5 & - & - &\textbf{74.1} &86.8 \\

    TRX~\cite{perrett2021temporal} & ResNet-50 & - & 96.1 & - & 75.6 & - & 64.6 & - &85.9 \\

    STRM~\cite{thatipelli2022spatio} & ResNet-50 & - & \textbf{96.9} & - & 77.3 & - &68.1 & - &86.7 \\
    
    \midrule
    \textbf{SA-CT w/o TMixer (ours)}& ResNet-50 & - & 96.4 & - & 77.8 & - & 61.4 & -&  86.4 \\
    \textbf{SA-CT (ours)} & ResNet-50 & \textbf{85.4} & 96.4 & \textbf{60.4} & \textbf{78.3}& \textbf{48.9} &\textbf{69.1} & 71.9 & \textbf{87.1} \\
    \textbf{SA-CT (ours)} & ViT-base & - & \textbf{98.0} & - &  \textbf{81.6} & - &  66.3 & - &  \textbf{91.2} \\
    
    \bottomrule   
    \end{tabular*}}
    \caption{\textbf{Comparisons with SOTAs.} 
    We hightlight that: 1) our ``SA-CT w/o TMixer'' that without any temporal information used outperforms the TRX and is comparable to the STRM on UCF101, HMDB51, and Kinetics. This reveals the importance of the spatial relation; 2) Our full SA-CT model improves the ``SA-CT w/o TMixer'' and achieves very competitive results. This indicates that the temporal information could be well utilized via a simple temporal module.
    }
    
     \vspace{-0.25in}
    \label{tab:compare_sota}
\end{table*}

\paragraph{\textbf{Implementation Details.}}
For fair comparison, following previous works~\cite{cao2020few, fu2020depth, zhu2018compound, perrett2021temporal}, we adopt the ResNet-50~\cite{he2016deep} pretrained on ImageNet~\cite{deng2009imagenet} as our backbone.
We remove the last two layers of ResNet-50 to extract feature maps of size $7\times 7\times 2048$ for video frames. The extracted feature maps can support our cross-attention operation on patches.
For each video (in both the query and support set), we uniformly sample 8 frames, i.e., $L=8$. We first re-scale the height of the frames to 256 and then crop the frames to $224\times 224$. Common data augmentations like random cropping and horizontal flipping are also used during model training.
The SGD is used as the optimizer for the meta-train stage. For UCF101, HMDB51, and Kinetics, the learning rate is set to 0.0005; while for SSv2, we use a learning rate of 0.005 due to its larger number of videos.
During training, the validate set is used to determine the hyper-parameters with the best are then adopted for testing.
For testing, we randomly select 10,000 meta-tasks from the test set and report the average accuracy as the final performance metric.

\subsection{Comparison with SOTAs}
\label{subsec:Compare with baselines}
To show the effectiveness of our SA-CT, we first compare our method with the most representative and SOTA methods, including the CMN~\cite{zhu2018compound}, 
TRAN~\cite{bishay2019tarn}, Embodied~\cite{fu2019embodied}, ARN~\cite{zhang2020few}, OTAM~\cite{cao2020few}, 
AmeFu-Net~\cite{fu2020depth}, 
TRX~\cite{perrett2021temporal}, and STRM~\cite{thatipelli2022spatio}. Our main results for 5-way 1-shot and 5-way 5-shot FSAR tasks are
reported in \cref{tab:compare_sota}.

\paragraph{\textbf{Results without the TMixer.}}
We first compare our SA-CT when the TMixer module is removed so as to verifies the importance of spatial relation (captured by the SCA module). 
As shown in "SA-CT w/o TMixer" in \cref{tab:compare_sota},
surprisingly, our SA-CT without utilizing any temporal information (i.e., ``SA-CT w/o TMixer (ours)'') without utilizing any temporal information achieves very competitive results to the current SOTAs on three out of four datasets.
Particularly, compared with TRX that only tackles the temporal misalignment, our ``SA-CT w/o TMixer" performs better on UCF101, HMDB51, and Kinetics datasets. This indicates that, on relatively common video datasets, aligning the spatial semantic concepts is more important than fixing the temporal inconsistencies. 
Even when compared with the STRM method which upgrades TRX by using the spatial information as a supplement, our results are still comparable on this three datasets. On HMDB51, we can even surpass STRM by 0.5\%.
Note that AmeFu-Net also reaches very competitive results, however, AmeFu-Net additionally introduces the depth modality while we only require the RGB frames. 
These results confirm the importance of modeling spatial relations for FSAR.
Our ``SA-CT w/o TMixer" does not work well on SSv2, which we conjecture is caused by the complicated temporal information contained in SSv2 videos. Arguably, such cases are relatively less common in real-world applications. The limitation on SSv2 in turn motivates us to incorporate the temporal information into our SA-CT by the TMixer to in turn boost the spatial relation.

\paragraph{\textbf{Results of the Full SA-CT}} 
Equipped with the TMixer, our full SA-CT model outperforms the baselines by a considerable margin consistently across different datasets and settings. 
Specifically, \textbf{1)} under the 5-way 1-shot setting, SA-CT achieves an accuracy of 85.4\%, 60.4\%, 48.9\%, and 71.9\% on the UCF101, HMDB51, SSv2, and Kinetics, respectively. 
We improve the ARN model by up to 19.1\%, 14.9\%, and 8.2\% respectively on UCF101, HMDB51, and Kinetics datasets. 
The superiority in such a low data regime indicates the efficiency of our SA-CT approach in capturing and aligning the spatial relations between the query and support videos.
\textbf{2)} Under the 5-way 5-shot setting, our SA-CT also achieves good results. On HMDB51, SSv2, and Kinetics, we outperform the best baseline STRM by 1.0\%, 1.0\%, and 0.4\%, respectively. 
Though the performance improvement may \textit{numerically} seems minor when compared with the improvement achieved under the 1-shot setting, we highlight that both TRX and STRM are well-developed and sophisticated FSAR algorithms that work really well under the 5-shot setting, and it is notably hard to exceed the two methods under this relatively data sufficient regime.
\textbf{3)} By employing the ViT-base architecture, we are able to achieve new state-of-the-art (SOTA) results on three benchmark datasets: UCF101, HMDB51, and Kinetics. The obtained accuracies for these datasets are 98.0\%, 81.6\%, and 91.2\%, respectively. For a comprehensive analysis of various backbones and their impact on performance, please refer to \cref{subsec:Other backbones}.

\subsection{Learning From LSPMs.}

\begin{table*}[]
    \centering
    \setlength{\tabcolsep}{8.5mm}\scalebox{0.8}{

    \begin{tabular}{cccccccc}
    \toprule   
    \textbf{Extractor} &\textbf{ Pretrain} & \textbf{Fix} & \textbf{Method} &  \textbf{UCF} & \textbf{HMDB} & \textbf{SSv2} & \textbf{Kinetics}   \\ 
    \midrule 

    ResNet-50 & SL/IN1K & \checkmark & PN-FSAR & 88.8 & 61.1 & 39.9 & 76.7 \\
    ResNet-50 & SL/IN1K & \checkmark & SA-CT & 94.2 & 73.7 & 59.5 & 85.3 \\
    \cellcolor{orange!15} ResNet-50 & \cellcolor{orange!15} SL/IN1K & \cellcolor{orange!15} - & 	\cellcolor{orange!15} SA-CT & 	\cellcolor{orange!15} \textbf{96.4} & 	\cellcolor{orange!15} 78.3 & 	\cellcolor{orange!15} \textbf{69.1} & 	\cellcolor{orange!15} 87.1 \\

    ResNet-50 & DINO/IN1K & \checkmark & PN-FSAR & 91.3 & 66.9 & 40.3 & 78.2 \\
    ResNet-50 & DINO/IN1K & \checkmark & SA-CT & 93.4 & 75.2 & 57.6 & 84.3 \\
    ResNet-50 & DINO/IN1K & - & SA-CT & 95.0 & 75.4 & 65.2 & 85.8 \\
    
    ResNet-50 & CLIP/YFCC & \checkmark & PN-FSAR & 91.3 & 70.6 & 36.5 & 82.2 \\
    ResNet-50 & CLIP/YFCC & \checkmark & SA-CT & 96.0 & \textbf{79.0} & 61.2 & \textbf{87.8} \\
    
    \midrule
    ViT-small & DINO/IN1K & \checkmark & PN-FSAR & 93.2 & 69.0 & 42.9 & 82.8 \\
    ViT-small & DINO/IN1K & \checkmark & SA-CT & 93.3 & 70.6 & 63.2 & 78.5 \\
    ViT-small & DINO/IN1K & - & SA-CT & 95.4 & 72.2 & 61.0 & 82.8 \\
    
    ViT-small & DeiT/IN1K & \checkmark & PN-FSAR & 91.8 & 64.2 & 36.8 & 82.7  \\
    ViT-small & DeiT/IN1K & \checkmark & SA-CT & 93.0 & 72.1 & 58.1 &83.7  \\
    ViT-small & DeiT/IN1K & - & SA-CT & \textbf{95.7} & \textbf{77.8} & \textbf{66.1} & \textbf{86.1}  \\
   
    \midrule
    ViT-base & SL/IN21K & \checkmark & PN-FSAR & 96.7 & 76.3 & 41.4 & 90.3 \\
     ViT-base & SL/IN21K & \checkmark & SA-CT & 96.3 & 77.6 & 51.8 & 88.6 \\
      \cellcolor{green!15}ViT-base & \cellcolor{green!15}SL/IN21K &  \cellcolor{green!15} - &  \cellcolor{green!15}SA-CT &  \cellcolor{green!15}\textbf{98.0} &  \cellcolor{green!15}\textbf{81.6} &  \cellcolor{green!15}\textbf{66.3} &  \cellcolor{green!15}\textbf{91.2} \\

    ViT-base & DINO/IN1K & \checkmark & PN-FSAR & 94.4 & 70.5 & 43.2 & 84.7 \\
    ViT-base & DINO/IN1K & \checkmark & SA-CT & 94.6 & 72.6 & 63.3 & 81.2 \\
    
    ViT-base & DeiT/IN1K & \checkmark & PN-FSAR & 92.8 & 67.0 & 35.2 & 82.0  \\
    ViT-base & DeiT/IN1K & \checkmark & SA-CT & 95.1 & 75.1 & 55.5 & 85.2  \\
    \bottomrule 
    \end{tabular}
    }
    \caption{\textbf{Learning from LSPMs.} 
    The 5-way 5-shot results are reported here. The ``PN-FSAR'' denotes the simple baseline that takes the non-parametric protonet as the FSL classifier. It basically evaluates the ability of LSPMs. Both the ResNet-50 and ViT are investigated with different pre-training methods. 
    Compared to our method (based on typical ResNet-50 (IN1K), highlighted in orange), a new SOTA (highlighted in green) is achieved by employing the ViT-base (ImageNet21k).} 
     \vspace{-0.25in}

    \label{tab:foundation}
\end{table*}

\label{subsec:Other backbones}
In this section, we investigate the benefit of using LSPMs for FSAR. Besides the ResNet-50 that has been widely used for FSAR~\cite{perrett2021temporal, thatipelli2022spatio, fu2020depth, cao2020few}, here we first explore different pretrained vision transformers as the feature extractors. 
We build our SA-CT method upon those extractors and use different LSPMs, including DINO~\cite{caron2021emerging}, CLIP~\cite{radford2021learning}, DeiT~\cite{touvron2021training}, and supervised training method (abbreviated as ``SL'') to initialize the backbones. We conduct experiments on both fixing and freeing the extractor parameters.
In addition, to evaluate the representations extracted by different extractors, we further introduce the ProtoNet (PN) ~\cite{snell2017prototypical} as a FSL classifier to build simple baselines. Concretely, PN is a non-prametric FSL method which classifies the query actions by ranking the feature similarities between instances. Thus, we form the ``PN-FSAR" by simply averaging the frames features and then feeding the averaged features into PN.
The 5-way 5-shot results are reported in Tab.~\ref{tab:foundation}.
From the results, we summarize the following observations:
\begin{itemize}
    \item \emph{A new SOTA
    (highlighted in green) 
    can be achieved by building our SA-CT upon the ViT-base (SL/IN21K)}. On UCF, HMDB, and Kinetics, the new SOTA accuracy reaches 98.0\%, 81.6\%, and 91.2\%, respectively. Compared to our prior SA-CT (ResNet-50)(SL/IN1K)
    which is highlighted in orange, the performances on UCF, HMDB, and Kinetics are improved by up to 1.6\%, 3.3\% and 4.1\%, respectively. These results confirm that FSAR task can indeed be improved by using LSPMs.
    
    \item \emph{ViT has advantages over the ResNet-50 on extracting richer representations for video frames}. Generally, with the same ``DINO/IN1K'' as pretraining, comparing the results of PN-FSAR, the ViT-base performs the best, followed by the ViT-small, while ResNet-50 performs the worst.
    
    \item \emph{Supervised training vs. self-supervised training}. Overall, though the new SOTA is achieved by the ViT-base pretrained under supervised learning (SL), we notice that the self-supervised learning (SSL) pretraining methods (e.g. DINO) also achieves good results. Specifically, taking ResNet-50 as the extractor, the SSL based ``DINO/IN1K'' is competitive to ``SL/IN1K''. For the ResNet-50 with PN-FSAR, the ``DINO/IN1K'' even outperforms the ``SL/IN1K''.
    
    \item \emph{Finetuning the extractor improves performance}. Comparing the results of fixing the extractor or not, we find that finetuning plays an important role in improving FSAR. Considering the potential data shifts between the pretraining and testing datasets, such an improvement is somewhat expected. That is, finetuning the base model on the specific testing data can help alleviate distributional shifts and improve performance.
    
    \item \emph{SSv2 is an exception}. Interestingly, among all the testing datasets, the UCF, HMDB, and Kinetics all benefit from LSPMs more or less, while SSv2 is the only exception. This reflects different properties of the testing datasets, i.e., compared to other datasets, SSv2 has a higher demand for temporal information~\cite{sevilla2021only}. Unfortunately, the LSPMs were all pretrained on images thus become less effective on datasets like SSv2. This calls for dedicated LSPMs for video tasks.
    
\end{itemize}

\begin{figure}[]
 \vspace{-0.15in}
    \centering    \includegraphics[width=1.\columnwidth]{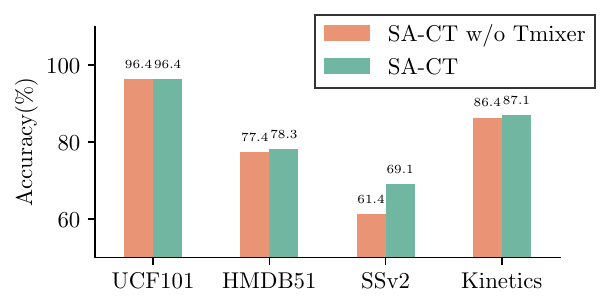}
    \vspace{-0.25in}
    \caption{\textbf{The impact of TMixer.} 
    The simple TMixer module improves the SA-CT steadily. }
    \vspace{-0.15in}
\label{without_temporal}
\label{fig:without_temporal}
\end{figure}

\begin{figure*}[h]
    \centering
    \includegraphics[width=0.8\linewidth]{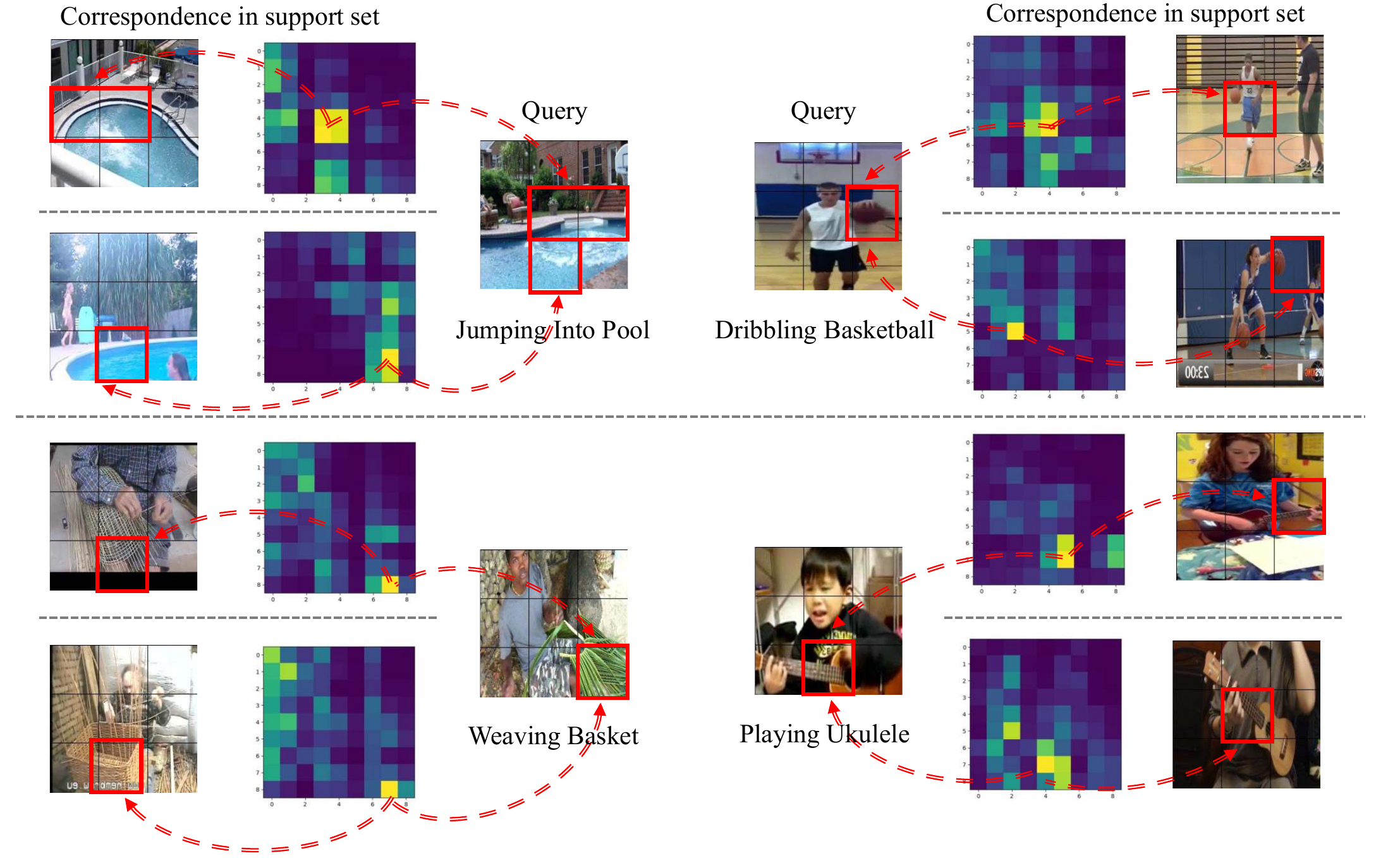}
    \vspace{-0.2in}
    \caption{\textbf{Visualizations of our spatial alignment.} 
    Four examples are given. The horizontal and vertical axes represent the support and query patches, respectively. The
    brighter the color, the higher the similarity. 
    We downsample the total patches from $7\times 7 = 49$ patches to $3\times 3 = 9$ patches for better visualization. Results show that our SA-CT successfully aligns the related regions of the support and query videos.}
    \label{visualization}
  \label{fig:visualization}
\end{figure*}

\begin{figure}[h]
    \centering
     \vspace{-0.1in}
    \includegraphics[width=0.8\columnwidth]{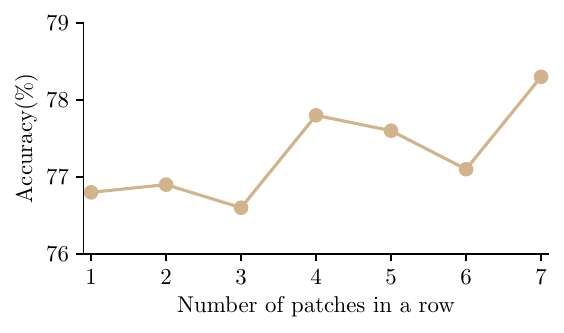}
     \vspace{-0.15in}
    \caption{
    \textbf{Ablation study on the number of patches.} We conduct experiments on HMDB51 with the number of patches in a row set varying from 1 to 7. SA-CT achieves the best result when the frames are split into $7 \times 7=49$ patches.
    }
     \vspace{-0.15in}
    \label{patches}
  \label{fig:patches}
\end{figure}

\subsection{Ablation Study}
\label{subsec:Ablation Study}

\paragraph{\textbf{The impact of TMixer.}} 
In ~\cref{fig:without_temporal},
we demonstrate that the TMixer module significantly improves SA-CT performance. On HMDB51 and Kinetics, accuracy increases from 77.8\% and 78.3\% to 87.1\% and 87.1\%, respectively. On SSv2, TMixer improves the performance from 61.4\% to 69.1\%, with a 7.7\% gain. These results confirm the effective utilization of temporal information using a simple MLP-based module.
TMixer also adds MLP3 and MLP4, reducing video frames from 8 to 4. This decreases the total Multi-Adds of SAC by up to 38.5\% (from 5.48G to 3.37G), as shown in ~\cref{tab:para_ablation}. The additional two MLPs contribute only 419.64M Multi-Adds. This demonstrates that TMixer is a simple and efficient module for utilizing temporal information and reducing computational costs.

 \begin{table}[h]
    \centering
    \scalebox{0.87}{
    \begin{tabular}{ccc}
    \toprule    
    \textbf{Frames} & \textbf{Total Multi-Adds of SAC} & \textbf{Total Multi-Adds of TMixer}   \\ 
    \midrule 
    8 & 5.48G & 419.64M \\
    4 & 3.37G & 839.28M \\
    \bottomrule    
    \end{tabular}
    }
    \caption{\textbf{TMixer accelerates the model.} 
    The Multi-Adds of SAC is decreased by 38.5\% (from 5.48G to 3.37G).
    }
     \vspace{-0.35in}
    \label{tab:para_ablation}
\end{table}

\paragraph{\textbf{Varying the number of patches.}}
Here we study the impact of varying number of patches on HMDB51 dataset.
In ~\cref{fig:patches}, the horizontal axis represents the number of patches in each row (e.g. 7 means each row has 7 patches, and one video frame has $7 \times 7=49$ patches), while the vertical axis represents the accuracy of the model. 
Generally, we observe that the accuracy of the performance improves as the number of patches increases with exceptions on 3 patches and 6 patches.
Critically, our SA-CT achieves the best performance when the number of patches is set as 7; and the performances are relatively poor when the number of patches is smaller than 4. This indicates that with more fine-grained patches, the benefits of our SCA module are maximized.

     \vspace{-0.1in}

\subsection{Visualization}
\label{subsec:Visualization}
In this section, we visualize four examples of correcting the spatial misalignment in ~\cref{fig:visualization} to help understand the working mechanism of our SA-CT.
Each example consists of a query frame image and a few support frame images, and the corresponding attention maps.

We downsample the total patches from $7\times 7=49$ patches to $3\times 3=9$ patches for better visualization. 
The vertical axis represent the query patches ($3\times 3$ grid, flatten to 9 patches), and  horizontal axis represent support patches ($3\times 3$ grid, flatten to 9 patches). The brighter the color, the higher the similarity.
The visualized examples show clearly the advantage of our SA-CT in aligning the objects of interest from different locations.
Taking ``playing ukulele'' as an example, our model successfully matches the ``ukulele'' patch even when it appears at different spatial locations of the two videos.
These results well indicate the effectiveness of our SCA module.

\section{Conclusion}
\label{sec:conclusion}
In this paper, we re-examined the role of spatial relations for Few-Shot Action Recognition (FSAR), and experimentally revealed its importance for accurate FSAR via a proposed Spatial Cross-Attention (SCA) module. 
With SCA, we introduced a novel \textbf{S}patial \textbf{A}lignment \textbf{C}ross \textbf{T}ransformer (\textbf{SA-CT}) which better handles the spatial misalignment by re-adjusting the spatial relations between two videos. Further equipped with a simple but effective Temporal Mixer module, our SA-CT achieves state-of-the-art results across different benchmark datasets and settings. We also conducted extensive experiments to explore the potential of LSPMs for FSAR. Our work contributes to the field with a new SOTA method and useful understanding for future research. 

\begin{acks}
  This project is in part supported by NSFC under Grant No. 62032006 and No. 62072116.
\end{acks}

\bibliographystyle{ACM-Reference-Format}
\bibliography{mm2023-cameraready}

\end{document}